\documentclass[10pt]{article}
\usepackage{graphicx} 
\usepackage[margin=1in]{geometry}
\usepackage{titlesec}
\usepackage{multicol}
\usepackage{times}
\usepackage{amsmath}
\usepackage{float}
\usepackage{dblfloatfix}
\usepackage[colorlinks=true,linkcolor=black,anchorcolor=black,citecolor=black,filecolor=black,menucolor=black,runcolor=black,urlcolor=black]{hyperref}
\usepackage{caption}
\usepackage{subcaption}
\usepackage{xcolor}
\usepackage[export]{adjustbox}
\usepackage{xurl}
\usepackage{fancyhdr}
\usepackage{hanging}

\makeatletter
\newcommand\notsotiny{\@setfontsize\notsotiny{6.5}{7.3}}
\makeatother

\titleformat*{\section}{\normalsize\bfseries}
\titleformat*{\subsection}{\normalsize\itshape}
\titleformat*{\subsubsection}{\normalsize\itshape}
\titleformat*{\paragraph}{\normalsize\itshape}
\titleformat*{\subparagraph}{\normalsize\itshape}

\titlespacing\section{0pt}{8pt plus 0pt minus 0pt}{2pt plus 0pt minus 0pt}
\titlespacing\subsection{0pt}{8pt plus 0pt minus 0pt}{2pt plus 0pt minus 0pt}
\titlespacing\subsubsection{0pt}{8pt plus 0pt minus 0pt}{2pt plus 0pt minus 0pt}

\begin{document}

\pagestyle{fancy}
\fancyhead{}
\renewcommand{\headrulewidth}{0pt}
\fancyfoot[L]{}
\fancyfoot[C]{}
\fancyfoot[R]{\thepage}

\begin{center}
    \textbf{Automatic Data Processing for Space Robotics Machine Learning}
    ~\\ ~\\
    \textbf{Anja Sheppard$^\text{a}$ and Katherine A. Skinner$^\text{a}$}
\end{center}

\noindent $^\text{a}$ \textit{Department of Robotics, University of Michigan, 2505 Hayward St, Ann Arbor, MI 48109 USA,} \{anjashep, kskin\} at umich.edu

\begin{center} \textbf{Abstract} \end{center}
\vspace{-9pt}
\hspace*{\parindent} Autonomous terrain classification is an important problem in planetary navigation, whether the goal is to identify scientific sites of interest or to traverse treacherous areas safely. Past Martian rovers have relied on human operators to manually identify a navigable path from transmitted imagery. Our goals on Mars in the next few decades will eventually require rovers that can autonomously move farther, faster, and through more dangerous landscapes--demonstrating a need for improved terrain classification for traversability. Autonomous navigation through extreme environments will enable the search for water on the Moon and Mars as well as preparations for human habitats. Advancements in machine learning techniques have demonstrated potential to improve terrain classification capabilities for ground vehicles on Earth. However, classification results for space applications are limited by the availability of training data suitable for supervised learning methods. This paper contributes an open source automatic data processing pipeline that uses camera geometry to co-locate Curiosity and Perseverance Mastcam image products with Mars overhead maps via ray projection over a terrain model. In future work, this automated data processing pipeline will be leveraged for development of machine learning methods for terrain classification. \\

\noindent \textbf{Keywords:} robotics, computer vision, geographic information systems, open source, space

\begin{multicols}{2}

\section*{Nomenclature}

\noindent $\varepsilon{}_{\text{rover}}$: rover mast elevation angle

\noindent $\alpha_{\text{rover}}$: rover mast azimuth angle

\noindent $\mathcal{E}$: absolute elevation angle range

\noindent $\mathcal{A}$: absolute azimuth angle range

\section*{Abbreviations}

\noindent Digital Elevation Model (DEM)

\noindent Geographic Information System (GIS)

\noindent High Resolution Imaging Science Experiment (HiRISE)

\noindent Parameter Value Language (PVL)

\noindent Planetary Data Reader (PDR)

\noindent Planetary Data System (PDS)

\noindent Position Localization and Attitude Correction Estimation Storage (PLACES)

\noindent Quantum Geographic Information System (QGIS) 

\noindent Rover Motion Counter (RMC)

\;

\section{Introduction}
\vspace{-3pt}

\begin{figure}[H]
    \centering
    \includegraphics[width=0.5\textwidth]{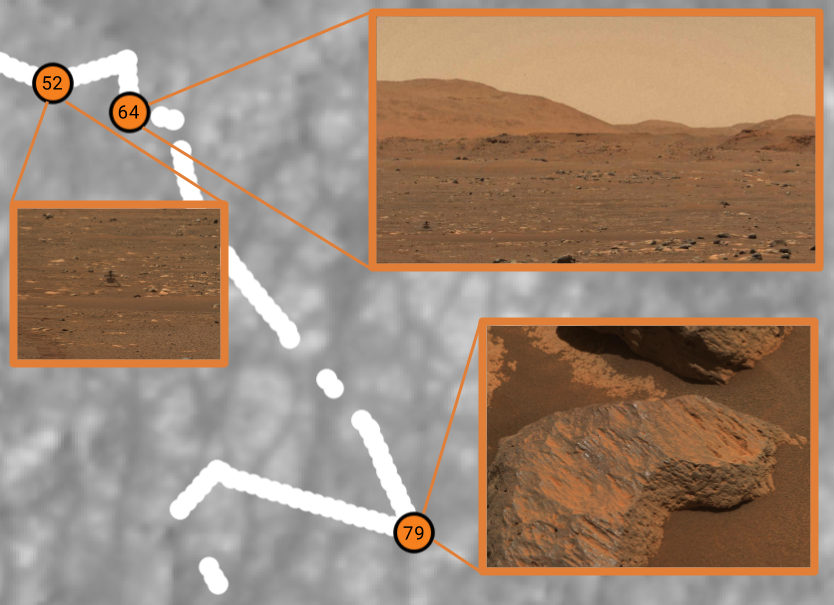}
    \caption{Co-location of Perseverance MastcamZ images \cite{mars2020data} with a HiRISE aerial map \cite{mars2020placesdata} on sols 52, 64, and 79.}
    \label{fig:co-location}
\end{figure}

Planned robotic exploration of Mars in the coming decades \cite{colaprete, vago, gibney} will need to be equipped with advanced autonomous capabilities to make navigational decisions and assess risky terrain. Learning-based approaches to perception \cite{zamanakos}, planning \cite{okereke}, and control \cite{liu} are enabling rapid improvements for Earth-based applications in robotics. In particular, self-supervised learning from multiple data modalities for terrain classification \cite{kurobe} and few-shot learning for autonomous navigation of mobile robotic platforms \cite{luong} are pushing the limits of how much can be learned from a limited or domain-specific dataset here on Earth. NASA also has a stake in the race to use artificial intelligence in space: in a whitepaper submitted to the 2020 Decadal Survey, scientists strongly emphasized the growing importance of machine learning in planetary science \cite{azari}. However, NASA's open source imagery is cataloged in a format unsuitable for machine learning applications, organized by date of capture rather than features, and lacks cross-referencing capabilities among different data products. This makes it challenging for non-expert users to locate images in specific locations, with specific features, or in relation to other instrument data--such as the matching between orbital and ground imagery seen in Fig. \ref{fig:co-location}. 

In this paper, we present an automatic data processing pipeline to co-locate NASA Curiosity and Perseverance Mastcam images with georeferenced aerial maps in the popular open source geographic data management platform Quantum Geographic Information System (QGIS). Our approach relies only on open source data and software, and it aims to improve data interoperability by creating a framework for synthesizing space sensory data for easier input into machine learning models.

This paper is organized as follows: in Section 2 is an overview of previous work in open source space data pipelines. Section 3 details the individual components of the proposed framework for publicly available data collected from both the Curiosity and Perseverance rovers, and Section 4 presents qualitative validation of the co-location results. In Section 5, we discuss limitations of this work and future directions. Open source code is available at \small{\url{https://github.com/umfieldrobotics/Automatic-Data-Processing-for-Space-Robotics-Machine-Learning}}.

\normalsize
\section{Background}

\subsection{Machine Learning in Space}

Several open source space datasets for machine learning have been developed in the past decade. One such dataset is LandCoverNet, a dataset of 8,941 labelled images of land cover from Sentinel-2 multi-spectral satellite images \cite{Alemohammad2020}. AstroVision, a dataset published just last year, contains 115,970 densely-annotated images of small bodies captured by various spacecraft \cite{driver}. NASA has developed the AI4MARS dataset, which contains 35,000 labelled terrain images from the Curiosity, Spirit, and Opportunity rovers \cite{swan}. While these datasets provide high-quality open source labelled datasets for space applications, there are still many domains that continue relying on raw data products from NASA's Planetary Data System (PDS) for learning applications. While software tools exist for accessing spacecraft data, such as NASA's SPICE \cite{spice1, spice2}, there are few publicly-available methods for integrating data across instruments for machine learning applications other than manually cross-referencing label files.

\subsection{NASA Planetary Data System}

All publicly released instrument data from NASA orbiters and rovers are stored on PDS for open access. For missions to Mars, there are several ways to view data: the PDS Image Atlas, the Mars Orbital Data Explorer \cite{wang}, the Planetary Image Locator Tool \cite{bailen}, and more. Each image product is paired with a text label (either in \texttt{.txt}, \texttt{.json}, or \texttt{.xml} formats) which contains detailed state information at the time of image capture. These tools operate separately: ground imagery is acquired through the PDS Image Atlas and orbital imagery is distributed through the Planetary Image Locator Tool--with no supported open source interface between the two. Co-locating images via geographic state information in their label files is one solution to this challenge.

\subsection{Co-location and Aerial-to-Ground Image Synthesis}

Image co-location, also called aerial-to-ground or satellite-to-ground image synthesis, refers to the problem of determining the true alignment between overhead images from satellites and ground images from a ground station, as seen in Fig. \ref{fig:co-location}. There is a growing interest in developing machine learning models for image co-location from distinguishable features in ground images \cite{regmi, workman, lu}. For network training and benchmarking, this recent work relies heavily on datasets that exploit the camera geometry from Earth imagery satellites and Google Street View to match corresponding viewsheds \cite{workman, lin, workman2015}.  Other works \cite{shokr, nachon} have both demonstrated methods for synthesizing satellite and ground imagery, but due to the unique camera geometries in each application, each approach tends to be specialized to the camera system. This paper's approach follows the lineage of camera geometry-based solutions by using a ray projection model over an elevation map to retrospectively determine what aerial area was covered in a single rover ground image.

On Earth, the Moon, and Mars, we have large amounts of satellite coverage, and so developing better methods of understanding embedded information in aerial images from corresponding ground images could improve the utility of orbital imagery in areas where ground images are not available. There are no open source datasets that provide aerial-to-ground synthesis for Mars. However, prior work in \cite{nachon} has laid the groundwork for this through proposing a method for co-location using the Curiosity rover camera geometry in a proprietary software. Our paper builds on \cite{nachon}'s work by extending their methods to the Perseverance rover, building the camera geometry co-location on the QGIS open source platform, and automating the process through a Python implementation.

\subsection{Geographic Information System}

GIS is a well-used framework for integrating geographically referenced information. There are two common platforms: ArcGIS, a paid software product, and QGIS, an open source alternative \cite{QGIS_software}. QGIS can be used with both a graphical interface and a Python module called PyQGIS. ArcGIS is the chosen platform for \cite{nachon}'s work, where they manually create \texttt{.csv} files containing Mastcam azimuth and elevation information for input into proprietary viewshed calculation tools. Our approach improves upon this previous work by automating the extraction of Mastcam positioning information from PDS image labels, conducting geometry calculations in Python, and generating viewsheds with open source QGIS tools.

\begin{figure*}
    \centering
    \includegraphics[width=0.96\textwidth]{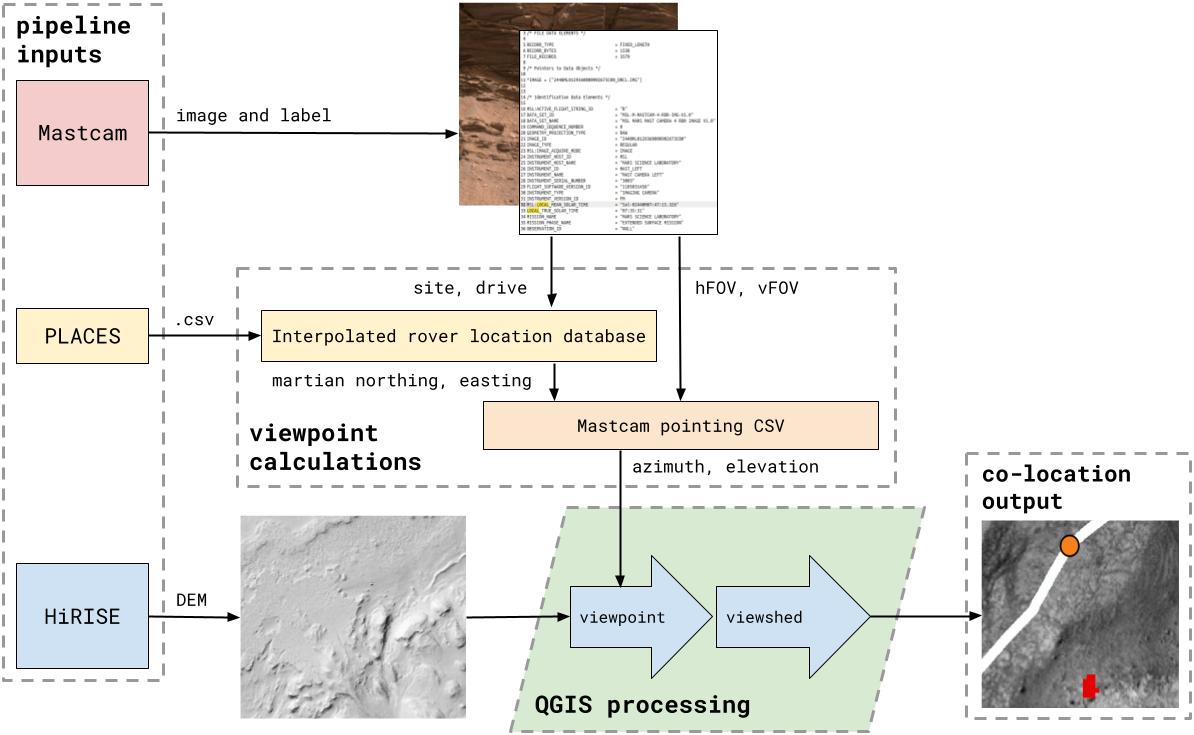}
    \caption{Flowchart overview of automatic aerial-to-ground synthesis pipeline. On the left, HiRISE \cite{msl_places}, PLACES, and Mastcam PDS \cite{msldata} data are utilized for viewpoint calculations. In the form of a \texttt{.csv} file, the viewpoint information is then processed by QGIS to produce a viewshed.}
    \label{fig:flowchart}
\end{figure*}

\section{Technical Approach}

The architecture of our co-location data processing pipeline is shown in Fig. \ref{fig:flowchart}. Our proposed pipeline contains three distinct components: the pipeline data input streams, the viewpoint calculations, and the QGIS processing.

\subsection{Pipeline Inputs}

\begin{figure*}[t]
    \centering
    \includegraphics[width=\textwidth]{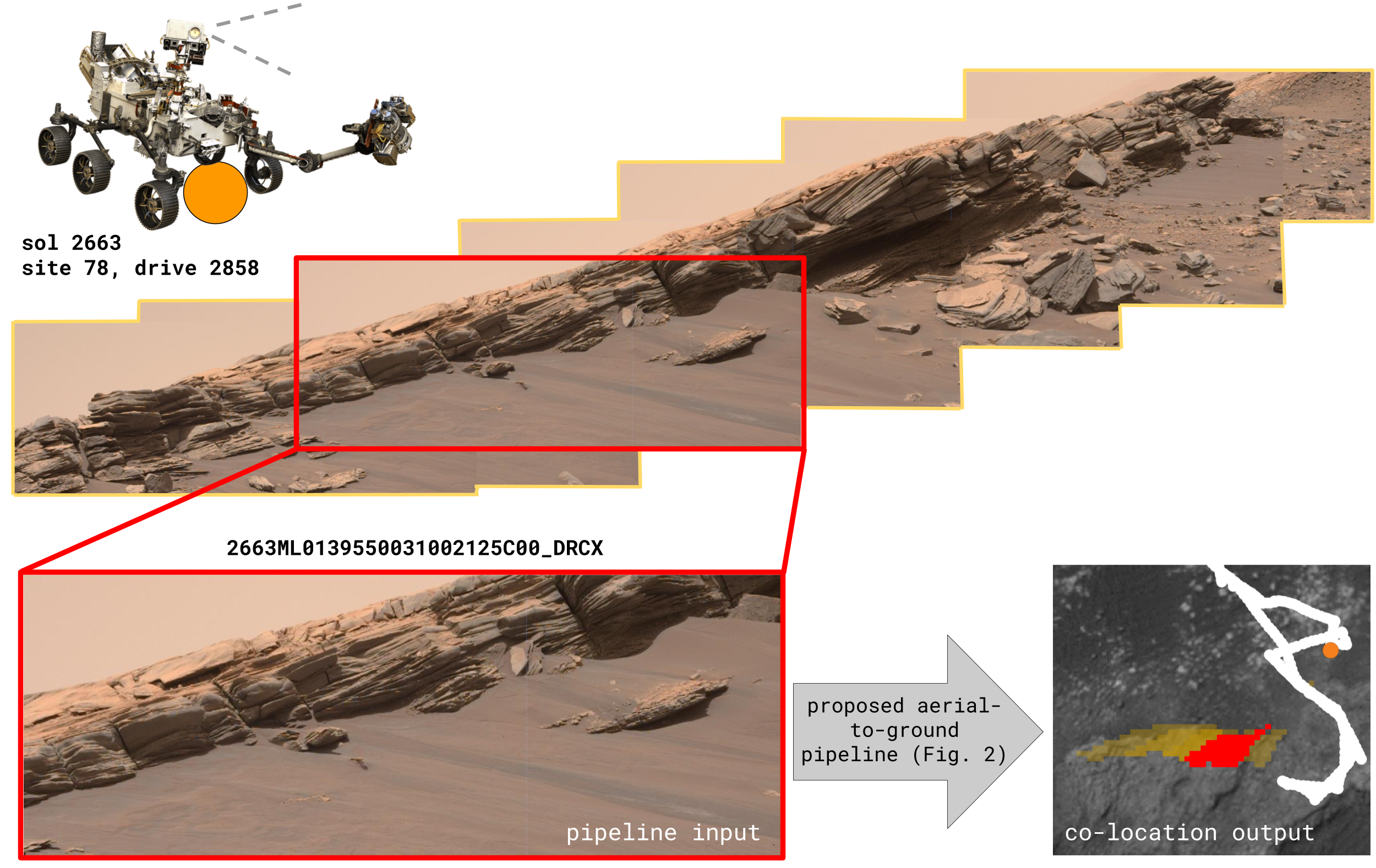}
    \caption{Overview of proposed aerial-to-ground synthesis pipeline (details in Fig. \ref{fig:flowchart}) input and output for an example Curiosity Mastcam image from sol 2663. Full image mosaic in yellow, with one component selected input image \texttt{2663ML0139550031002125C00\_DRCX} \cite{msldata} in red. The co-location output displays the viewsheds for the full mosaic to demonstrate how this tool provides geographic context for images. Full list of images used in the Appendix.}
    \label{fig:mosaic_flowchart}
\end{figure*}

\subsubsection{Mars Reconnaissance Orbiter HiRISE}

The High Resolution Imaging Science Experiment (HiRISE) has been imaging the surface of Mars at a resolution of 0.3 meters per pixel since the early 2000s \cite{ebben}. NASA then mosaics these images together into georeferenced maps and Digital Elevation Models (DEMs) of the Martian surface, released as a part of the Position Localization and Attitude Correction Estimation Storage (PLACES) bundle. The terrain information in the DEM is used during the viewshed calculation to determine what areas of the Martian surface were visible to the Mastcam at a given azimuth and elevation angle of the camera.

\subsubsection{Rover Camera Systems}
Our pipeline can also take data from rover camera systems as input. We focus on integrating camera data from two rover platforms: Curiosity and Perseverance. The Curiosity rover has several sets of engineering cameras: four pairs of fixed hazard avoidance cameras (Hazcams) for viewing terrain within 10 feet, two pairs of navigation cameras (Navcams) for black and white 3D terrain information, and one Mastcam mounted on a movable mast for color images and videos \cite{maki, malin}. The Perseverance rover also has a camera mounted on a mast--similarly named the MastcamZ for its ability to zoom \cite{bell}. Mastcam and MastcamZ images have varied objects of study, angles of focus, and vantage points of Martian landscapes--making them the most useful for understanding Martian terrain but also the most difficult to co-locate. The pipeline can take in any Mastcam or MastcamZ image and their corresponding label, but tends to perform best on the full-size radiometrically corrected version. This is indicated in the file name by the \texttt{DRCL} processing code for Curiosity images \cite{mslsis} and the \texttt{N} (non-thumbnail) identifier for Perseverance images \cite{mars2020sis}.

\subsubsection{Image Label Files}

Each rover platform outputs image label files in a distinct format. The Curiosity rover pairs a \texttt{.txt} label file in PDS3 format with each released image. These label files contain information describing the rover location, mast orientation, and camera field of view when the image was taken--all critical information for geographical localization. An example of relevant lines from the label file for image \texttt{1429MR0070680170702598E01\_DRCL} \cite{msldata} is below.

\begin{scriptsize}
\begin{verbatim}
ROVER_MOTION_COUNTER_NAME = ("SITE", "DRIVE", "POSE", 
                             "ARM", "CHIMRA", "DRILL", 
                             "RSM", "HGA", 
                             "DRT", "IC")
ROVER_MOTION_COUNTER      = (56, 1632, 
                             8, 0, 
                             0, 0, 
                             142, 90, 
                             0, 0 )
\end{verbatim}
\end{scriptsize}

In the Mars2020 mission, the Perseverance rover releases data product information in the \texttt{.xml} PDS4 format. An example excerpt from the label file corresponding to the image \texttt{ZLF\_0089\_074\_855262\_265RA\linebreak D\_N0040048ZCAM08050\_034085J03} \cite{mars2020data} is below.

\begin{notsotiny}
\begin{verbatim}
<geom:coordinate_space_frame_type>
    ROVER_NAV_FRAME
</geom:coordinate_space_frame_type>
<geom:Coordinate_Space_Index>
    <geom:index_id>SITE</geom:index_id>
    <geom:index_value_number>4</geom:index_value_number>
</geom:Coordinate_Space_Index>
<geom:Coordinate_Space_Index>
    <geom:index_id>DRIVE</geom:index_id>
    <geom:index_value_number>48</geom:index_value_number>
</geom:Coordinate_Space_Index>
\end{verbatim}
\end{notsotiny}

Both of these excerpts show the same information: the \texttt{SITE} and \texttt{DRIVE} indices of the rover when the image was captured.

\subsection{Viewpoint Calculations}

The next stage of the pipeline is the viewpoint calculation, which extracts relevant information from the image label file, determines the global location of the rover, and calculates the absolute azimuth and elevation angles of the Mastcam. For each Curiosity Mastcam image that goes through the proposed co-location pipeline, the Parameter Value Language (PVL) Python library \cite{beyer} reads the rover \texttt{SITE} and \texttt{DRIVE} indices in order to index into the \texttt{localized\_interp.csv} to retrieve the rover northing and easting position at the time the image was taken. \texttt{localized\_interp.csv} is released as a part of NASA PDS PLACES bundle \cite{mslplacesdata}, which is the best estimate of the rover's location. 

Similarly, Perseverance MastcamZ image's \texttt{SITE} and \texttt{DRIVE} indices are retrieved from the \texttt{.xml} PDS4 label file via the Planetary Data Reader (PDR) Python library \cite{million}. The Mars2020 mission PLACES bundle provides \texttt{best\_interp.csv} \cite{mars2020placesdata} as a reference for the rover location in global easting and northing coordinates. 

Both the Curiosity and the Perseverance define their location through the Rover Motion Counter (RMC), a set of indices that uniquely identifies every location along the rover's trajectory. The \texttt{SITE} index defines a local world frame, beginning with the landing site and increasing at the discretion of rover operators. \texttt{DRIVE} defines the position of the rover in relation to the site, and it increments whenever the rover moves. The \texttt{DRIVE} index is odd when the rover is moving, and even when it is stationary \cite{msl_places}. The image label also provides the Mastcam azimuth and elevation angles, which describe the heading of the rover camera in relation to the horizon and north when the image was captured. Elevation is defined with $0$ degrees when the Mastcam is pointed at the horizon, and positive when the camera rotates towards the sky. The azimuth is defined as clockwise positive, with $0$ degrees at north.

Using the mast fixed azimuth and elevation, we can calculate the absolute horizontal field of view ($\mathcal{A}_{\text{left}} \rightarrow \mathcal{A}_{\text{right}}$) for any given Mastcam or MastcamZ image through the following equations:
\begin{equation}
    \mathcal{A}_{\text{left}} = \alpha_{\text{rover}} - \frac{\text{hFOV}}{2}
    \label{eq:lhfov}
\end{equation}
\begin{equation}
    \mathcal{A}_{\text{right}} = \alpha_{\text{rover}} + \frac{\text{hFOV}}{2}
    \label{eq:rhfov}
\end{equation}
where $\alpha_{\text{rover}}$ is the fixed mast azimuth angle and the hFOV is the relative horizontal field of view from the label file. Additionally, the absolute vertical field of view ($\mathcal{E}_{\text{upper}} \rightarrow \mathcal{E}_{\text{lower}}$) is calculated by:
\begin{equation}
    \mathcal{E}_{\text{upper}} = \varepsilon_{\text{rover}} + \frac{\text{vFOV}}{2}
    \label{eq:uvfov}
\end{equation}
\begin{equation}
    \mathcal{E}_{\text{lower}} = \varepsilon_{\text{rover}} - \frac{\text{vFOV}}{2}
    \label{eq:lvfov}
\end{equation}
where $\varepsilon_{\text{rover}}$ is the fixed mast elevation angle and the vFOV is the relative vertical field of view from the label file. These values are written into a \texttt{.csv} file and then read into QGIS for further processing.


\subsection{QGIS Processing}

After the necessary components for calculating the viewshed are collected from the Mastcam image label and synthesized into a \texttt{.csv} file, the QGIS component of the pipeline comes into play. The Visibility Analysis \cite{cuckovic} plugin takes in a \texttt{.csv} input to establish the viewpoint of the rover when the image was taken. Once the viewpoint is created, Visibility Analysis calculates a viewshed, which projects rays along the bounds of the rover's field of view onto the DEM to discover the viewable area of the overhead map. This process is computationally intensive, and it takes on the order of tens of seconds to complete.

The result is a QGIS raster indicating which geographic area is is viewable from the viewpoint given the constraints of the Mastcam position and terrain occlusion. An overview of the pipeline inputs and outputs for a Curiosity Mastcam image mosaic is shown in Fig. 
\ref{fig:mosaic_flowchart}.

\section{Experiments \& Results}

\subsection{Curiosity}

There is no publicly-available ground truth for quantitative evaluation of the proposed co-location method. However, we can compare our results qualitatively to results from prior work and analyze overlapping viewsheds.

\begin{figure*}
     \centering
     \begin{subfigure}[b]{0.3\textwidth}
         \centering
         \includegraphics[width=\textwidth, height=\textwidth]{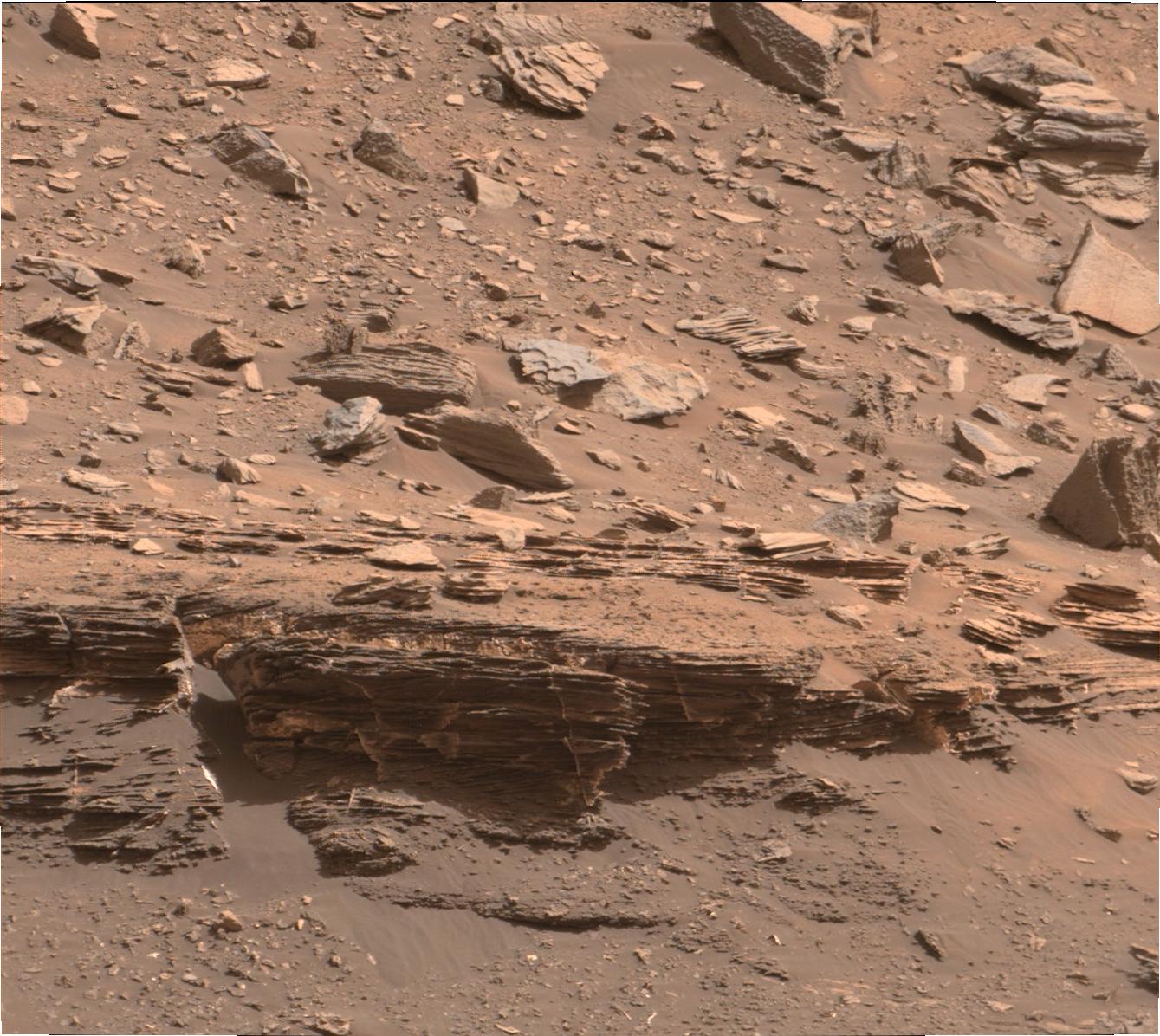}
        \caption{\texttt{1429MR0070680170702598E01}\cite{msldata}}\vspace{0.5em}
         \label{fig:orig_1}
     \end{subfigure}
     \hfill
     \begin{subfigure}[b]{0.3\textwidth}
         \centering
         \includegraphics[width=\textwidth, height=\textwidth]{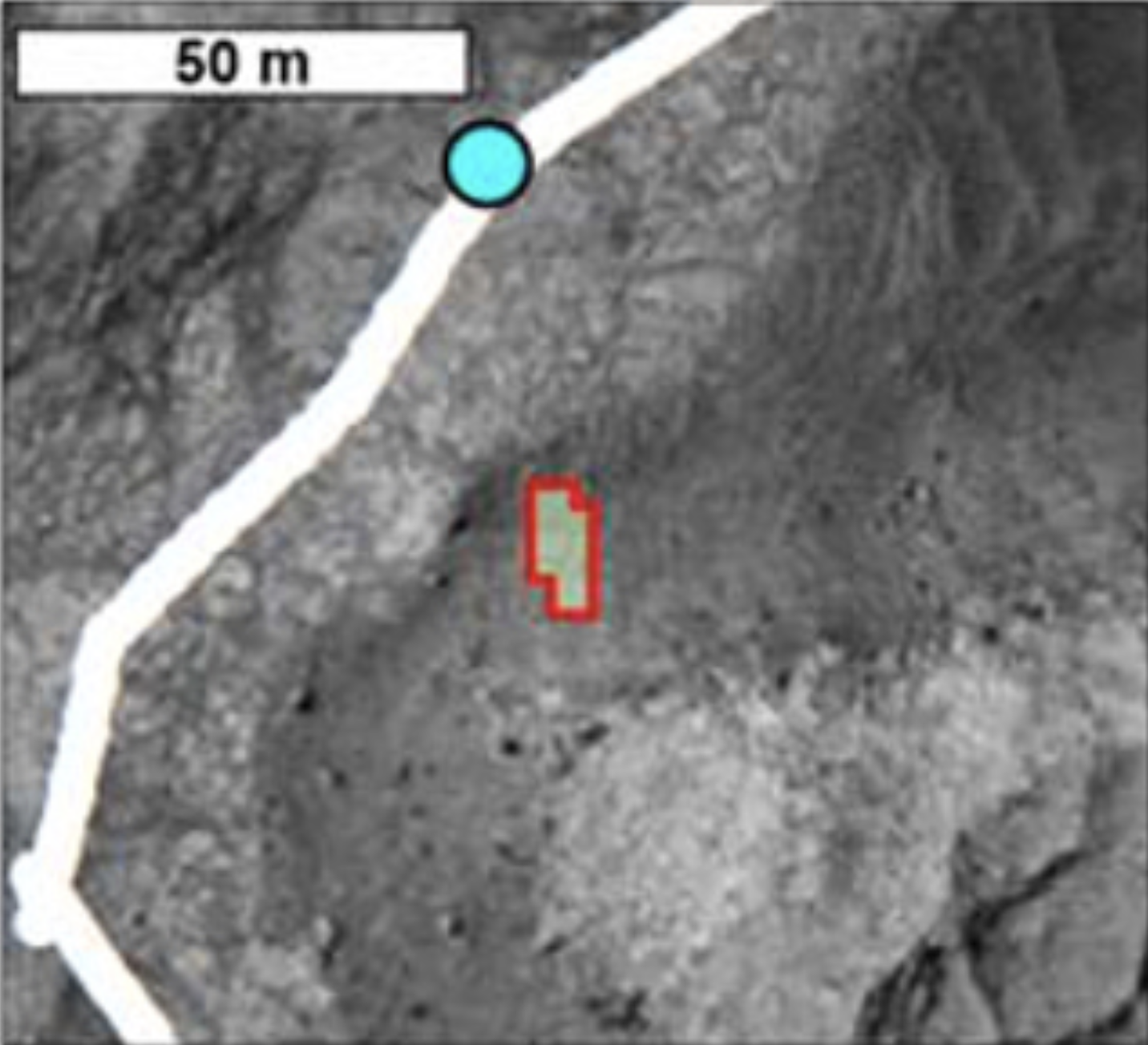}
         \caption{Nachon et. al. \cite{nachon}}\vspace{0.5em}
         \label{fig:nachon_1}
     \end{subfigure}
     \hfill
     \begin{subfigure}[b]{0.3\textwidth}
         \centering
         \includegraphics[width=\textwidth, height=\textwidth]{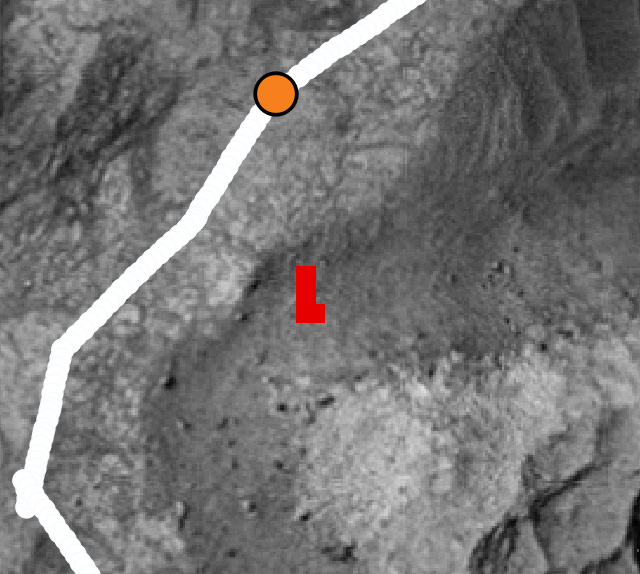}
         \caption{Ours}\vspace{0.5em}
         \label{fig:ours_1}
     \end{subfigure}
     \hfill 
     \\ [-3pt]
     \begin{subfigure}[b]{0.3\textwidth}
         \centering
         \includegraphics[width=\textwidth, height=\textwidth]{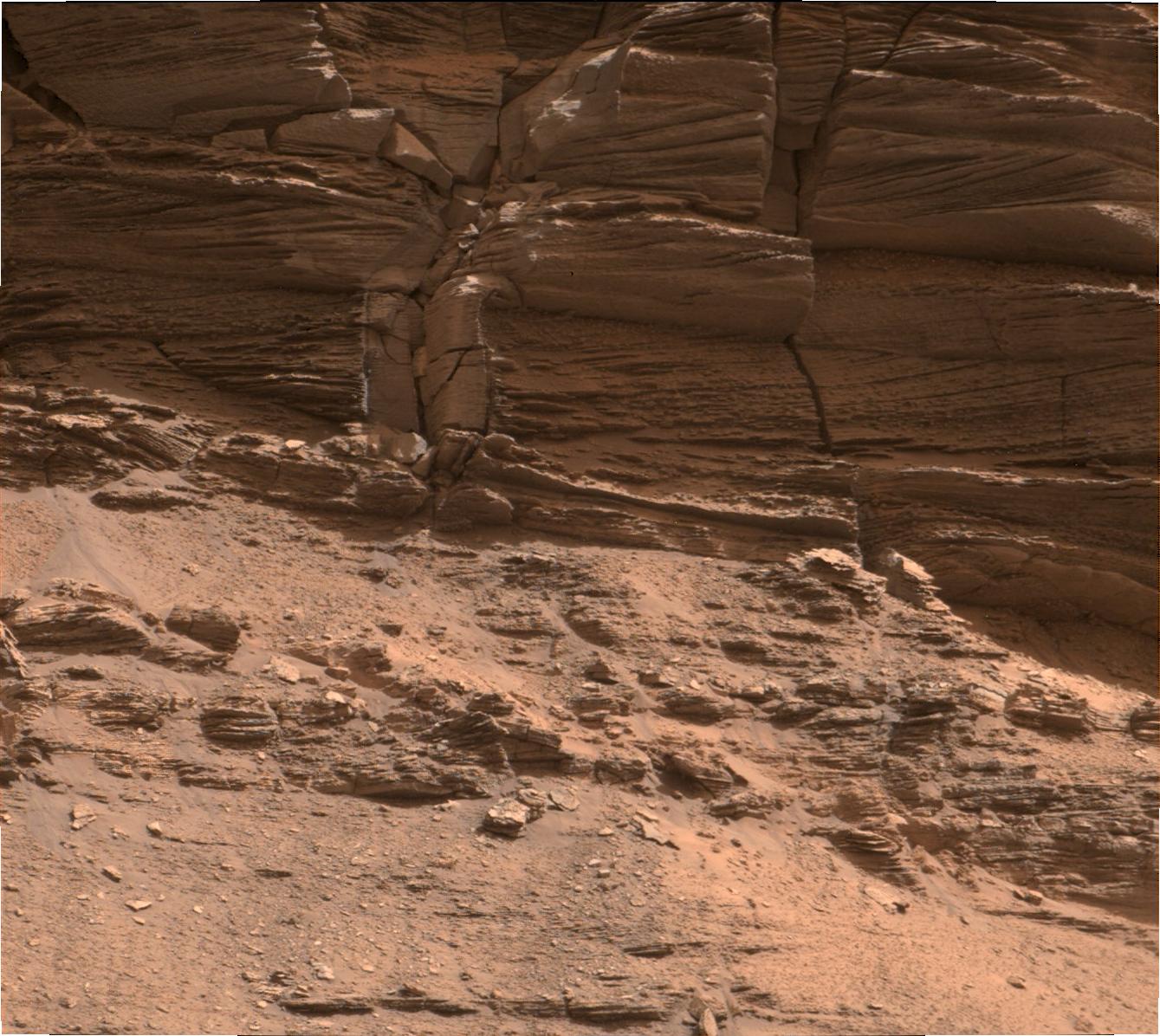}
         \caption{\texttt{1429MR0070680060702587E01}\cite{msldata}}\vspace{0.5em}
         \label{fig:orig_2}
     \end{subfigure}
     \hfill
     \begin{subfigure}[b]{0.3\textwidth}
         \centering
         \includegraphics[width=\textwidth, height=\textwidth]{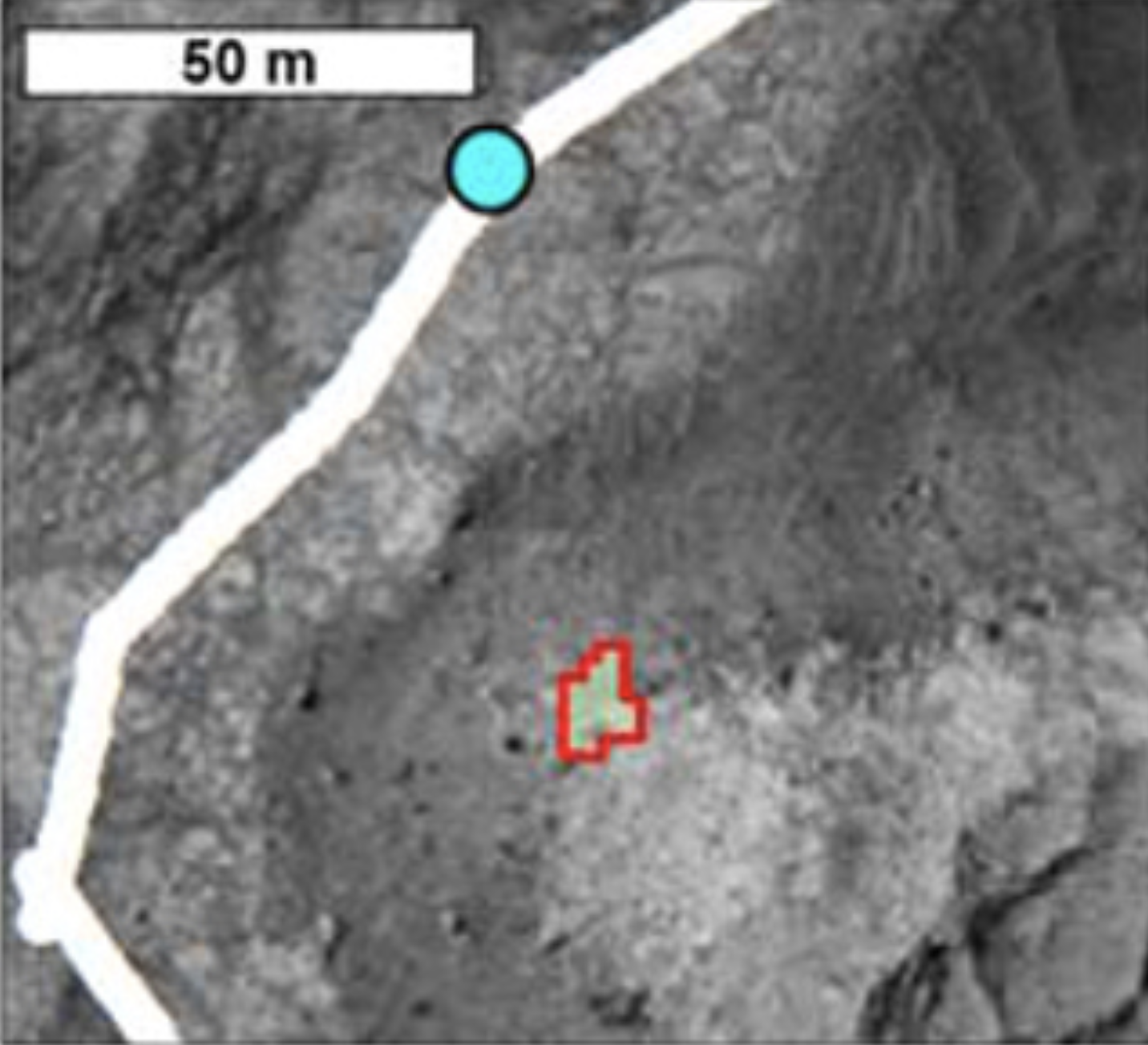}
         \caption{Nachon et. al. \cite{nachon}}\vspace{0.5em}
         \label{fig:nachon_2}
     \end{subfigure}
     \hfill
     \begin{subfigure}[b]{0.3\textwidth}
         \centering
         \includegraphics[width=\textwidth, height=\textwidth]{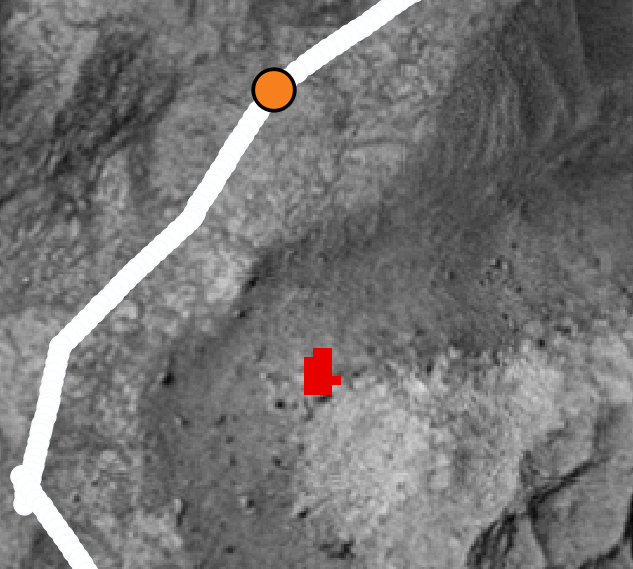}
         \caption{Ours}\vspace{0.5em}
         \label{fig:ours_2}
     \end{subfigure}
     \hfill 
     \\ [-3pt]
     \begin{subfigure}[b]{0.3\textwidth}
         \centering
         \includegraphics[width=\textwidth, height=\textwidth]{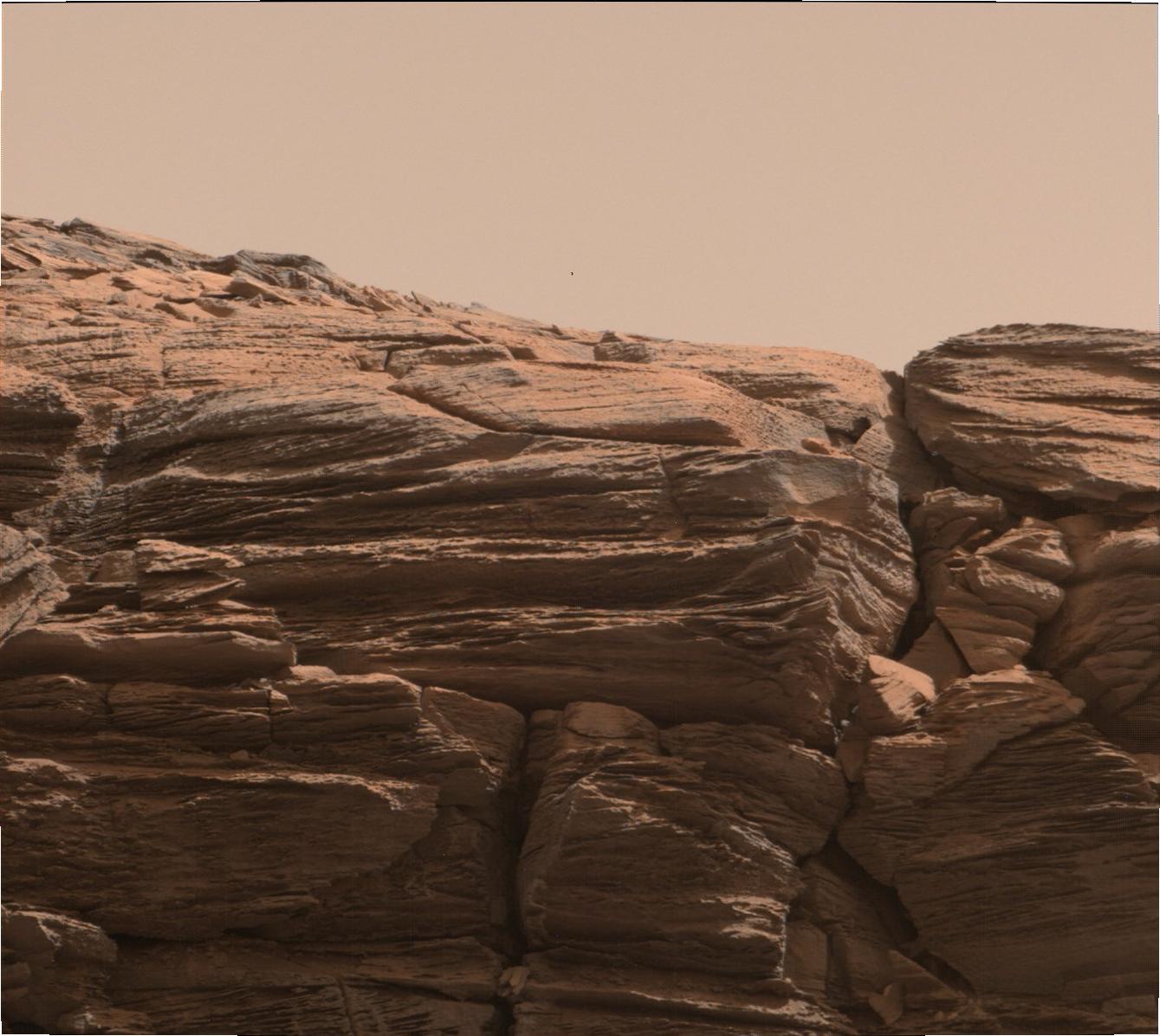}
         \caption{\texttt{1429MR0070680020702583E01}\cite{msldata}}\vspace{0.5em}
         \label{fig:orig_3}
     \end{subfigure}
     \hfill
     \begin{subfigure}[b]{0.3\textwidth}
         \centering
         \includegraphics[width=\textwidth, height=\textwidth]{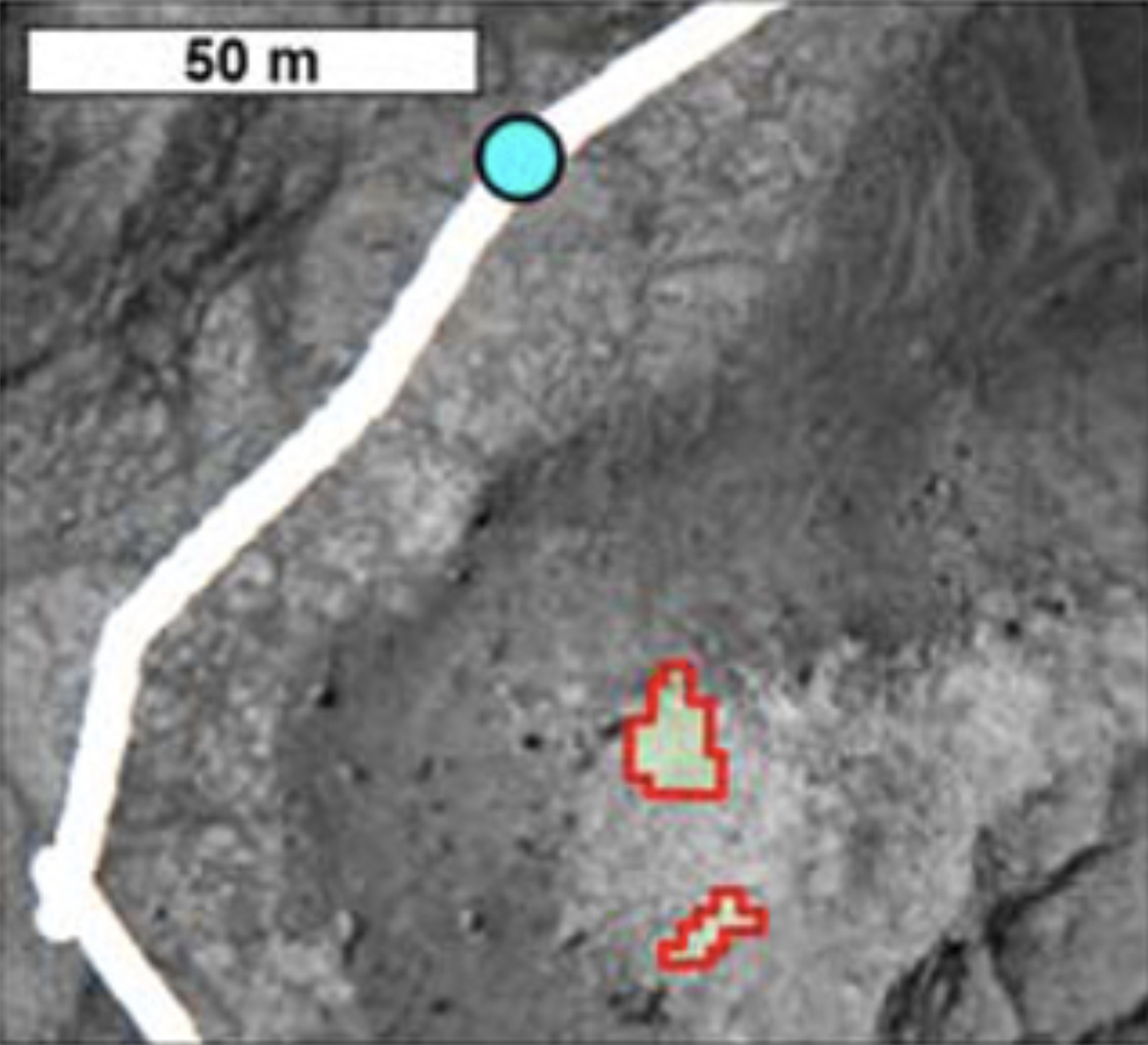}
         \caption{Nachon et. al. \cite{nachon}}\vspace{0.5em}
         \label{fig:nachon_3}
     \end{subfigure}
     \hfill
     \begin{subfigure}[b]{0.3\textwidth}
         \centering
         \includegraphics[width=\textwidth, height=\textwidth]{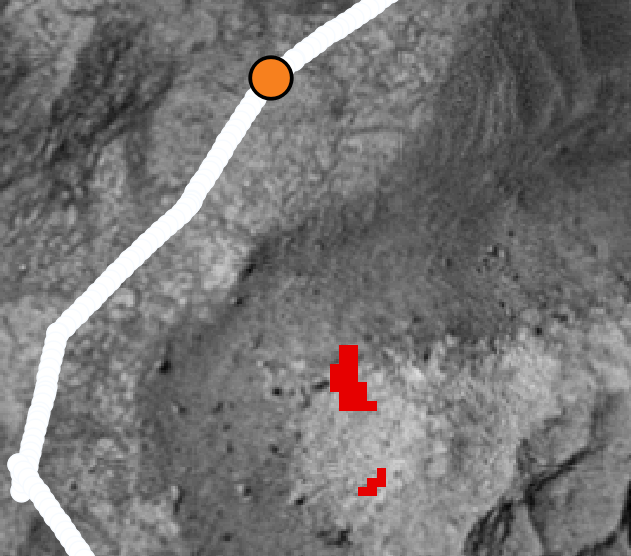}
         \caption{Ours}\vspace{0.5em}
         \label{fig:ours_3}
     \end{subfigure}
     \hfill 
     \\ [-3pt]
     \begin{subfigure}[b]{0.3\textwidth}
         \centering
         \includegraphics[width=\textwidth, height=\textwidth]{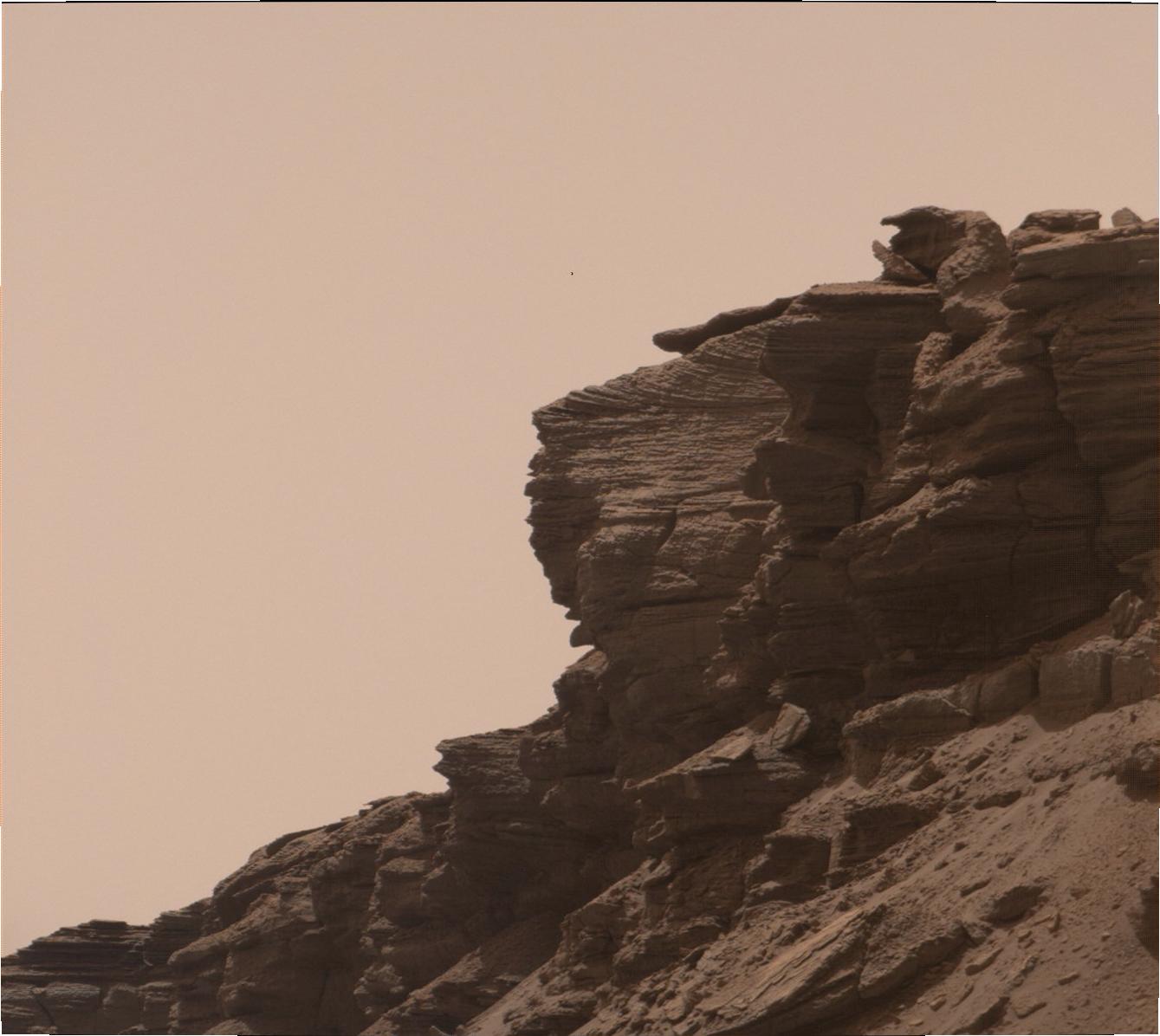}
         \caption{\texttt{1429MR0070670010702567E01}\cite{msldata}}\vspace{0.5em}
         \label{fig:orig_4}
     \end{subfigure}
     \hfill
     \begin{subfigure}[b]{0.3\textwidth}
         \centering
         \includegraphics[width=\textwidth, height=\textwidth]{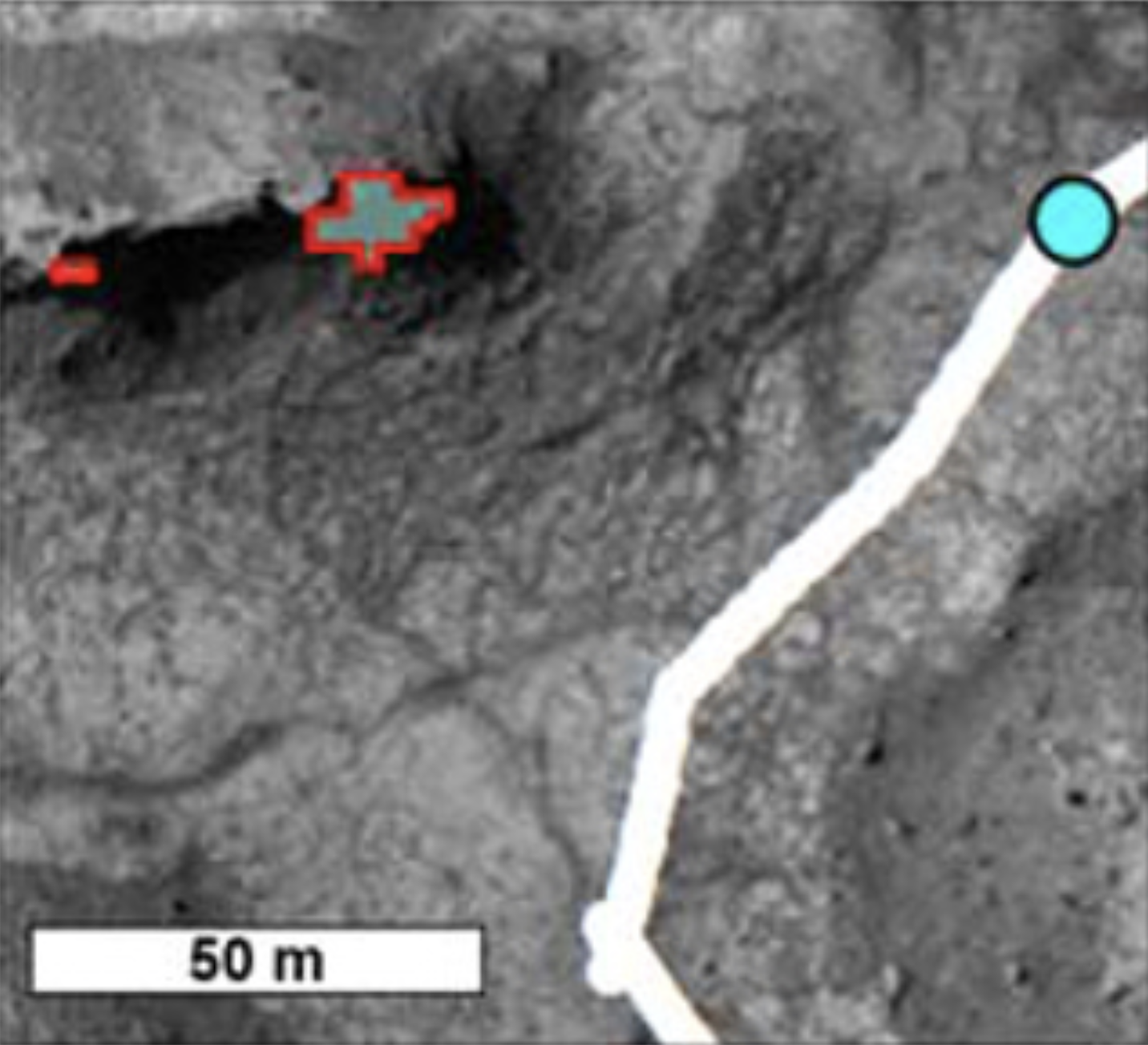}
         \caption{Nachon et. al. \cite{nachon}}\vspace{0.5em}
         \label{fig:nachon_4}
     \end{subfigure}
     \hfill
     \begin{subfigure}[b]{0.3\textwidth}
         \centering
         \includegraphics[width=\textwidth, height=\textwidth]{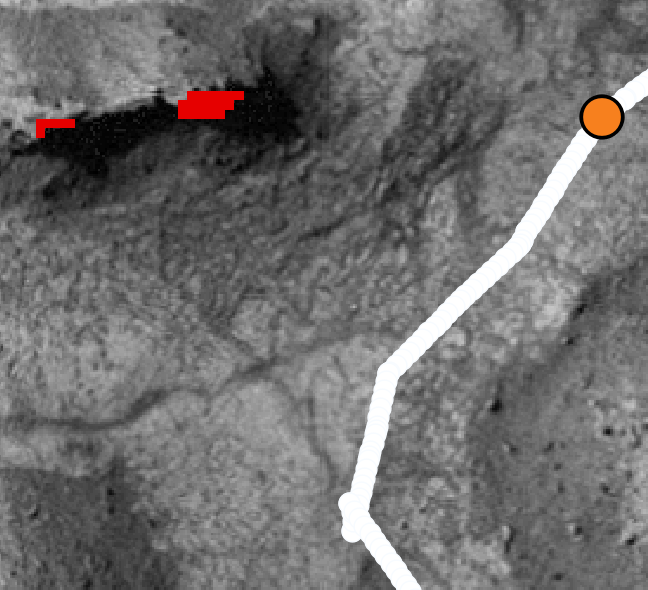}
         \caption{Ours}\vspace{0.5em}
         \label{fig:ours_4}
     \end{subfigure}
     \vspace{-12pt}
     \caption{\normalsize{Comparison of Curiosity co-location results from our method in QGIS (c, f, i, l) and the ArcGIS method of \cite{nachon} in ArcGIS (b, e, h, k). Images b, e, h, and k from \cite{nachon} are screenshotted from the paper.}}
     \label{fig:comparison}

\end{figure*}

Shown in Fig. \ref{fig:comparison} is the comparison between the co-location results of our method against the previous work of \cite{nachon}. Whenever there is a distinct foreground and background with occluded area in the Mastcam image, two distinct viewshed areas are shown in the co-located result. A visual analysis of the shapes, sizes, and locations of calculated viewsheds of \cite{nachon} and ours reveals comparable results. Our results are generated through an entirely open source autonomous pipeline, which opens this area up to future development and long-term maintenance.

\begin{figure}[H]
    \centering
    \includegraphics[width=.47\textwidth]{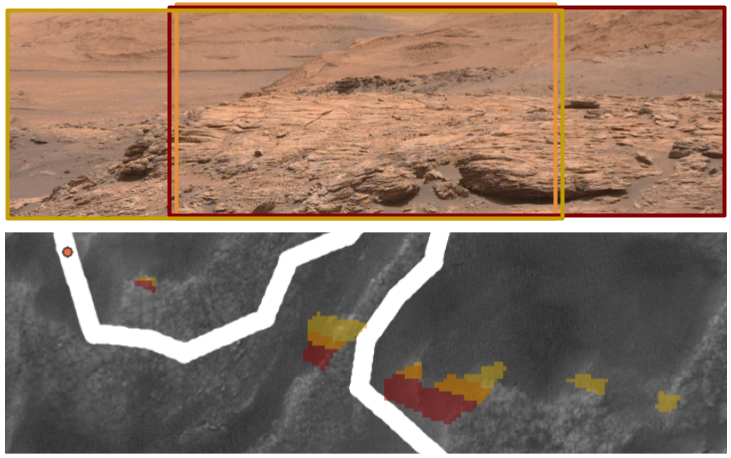}
    \caption{Overlapping viewsheds for overlapping images \texttt{2933ML0152980011102337C00\_DRCX} in yellow and \texttt{2933ML0152980021102338C00\_DRCX} \cite{msldata} in red from viewpoint in orange along rover path in white.}
    \label{fig:viewshed}
\end{figure}

Another approach to validate the results from this method involves analyzing the generated viewsheds from overlapping Mastcam images. If the Mastcam images are overlapping, then their calculated viewsheds should be as well. As seen in Fig. \ref{fig:viewshed}, two images are outlined in yellow and red, with their corresponding viewsheds in yellow and red as well. The overlapping components of their viewsheds are displayed in orange, which matches the expected size and area of overlap from visually inspecting the image mosaic.

\vspace{-2pt}

\captionsetup[figure]{font=small}

\captionsetup[figure]{font=normal}

\subsection{Perseverance}

\begin{figure}[H]
    \includegraphics[width=0.49\textwidth]{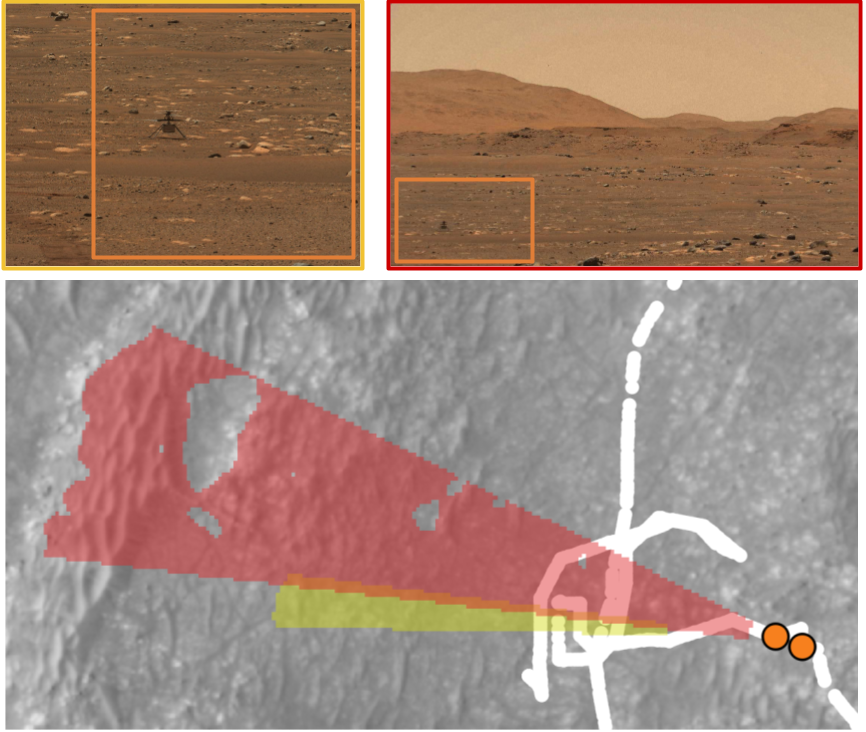}
    \caption{Overlapping Perseverance MastcamZ images and generated viewsheds from sols 52 (yellow, top left) and 64 (red, top right) \cite{mars2020data}. The image and viewshed overlapping area is shown in orange.}
    \label{fig:perseverance_overlap}
\end{figure}

As there is no previously published work that calculates the co-located viewsheds for the Perseverance rover, we take a similar approach to the previous section of analyzing the viewsheds from overlapping images to validate our results. Shown in Fig. \ref{fig:perseverance_overlap} are images \texttt{ZLF\_0064\_0672622163\_363RAF\_N0032046ZCAM0\linebreak5042\_026050J03} and \texttt{ZLF\_0052\_0671559321\_081\linebreak RAD\_N0031950ZCAM05024\_110085J01} from sols 64 and 52, respectively \cite{mars2020data}. Between sols 52 and 64, there were three Ingenuity flights that all lifted off and landed in the same location, so the location of Ingenuity is approximately static between those photos being taken. Additionally, there is a rock feature to the right of Ingenuity that is visible in both images. By observing these features in both images, we visually confirm that they have overlapping viewsheds despite being captured on different days. The resulting viewsheds from our autonomous pipeline agree with this conclusion.

\section{Conclusion}

NASA rovers produce an abundance of visual and geologic data. However, there are limited open source software tools available to enable machine learning integration for these data products. We demonstrate in this paper an automatic pipeline that can co-locate Mastcam and MastcamZ images with their respective georeferenced satellite maps. This allows for a greater understanding of the geographic context of each individual image and enables geographic features to play a larger role in selection of images for further study. Our pipeline is built with open source tools in order to ensure accessibility.

This paper lays the foundation for future work that synthesizes more sensor modalities, such as Curiosity's hands lens imager, x-ray spectrometer, and rock chemistry laser as well as Perseverance's subsurface radar, ultraviolet spectrometer, and weather station. We postulate that this work will be a valuable tool for increasing innovation and enabling open access to NASA rover data.

\section*{Acknowledgements}
This work was funded by the University of Michigan Robotics Department Fellowship. 

\section*{Appendix - Citation of PDS Data products}

\noindent PDS3 data products cited in this paper as part of \url{https://doi.org/10.17189/1520328} have the following PDS3 \texttt{DATA\_SET\_ID:PRODUCT\_IDs}:

\noindent \texttt{1429MR0070680170702598E01} \\
\texttt{1429MR0070680060702587E01} \\
\texttt{1429MR0070680020702583E01} \\
\texttt{1429MR0070670010702567E01} \\
\texttt{2663ML0139550021002124C00} \\
\texttt{2663ML0139550031002125C00} \\
\texttt{2663ML0139550041002126C00} \\
\texttt{2663ML0139550051002127C00} \\
\texttt{2663ML0139550061002128C00} \\
\texttt{2663ML0139550071002129C00} \\
\texttt{2663ML0139550081002130C00} \\
\texttt{2933ML0152980011102337C00} \\
\texttt{2933ML0152980021102338C00} \\

\noindent PDS4 data products cited in this paper as part of \url{https://doi.org/10.17189/bs6b-4782} include:

\begin{hangparas}{0.5in}{1}
\texttt{urn:nasa:pds:mars2020\_rover\_places:data\linebreak\_localizations:best\_interp.csv}

\texttt{urn:nasa:pds:mars2020\_mastcamz\_ops\_cali\linebreak brated:data:ZLF\_0052\_0671559321\_0\linebreak81RAD\_N0031950ZCAM05024\_110085J01}

\texttt{urn:nasa:pds:mars2020\_mastcamz\_ops\_cali\linebreak brated:data:ZLF\_0064\_0672622163\_3\linebreak63RAF\_N0032046ZCAM05042\_026050J03}

\texttt{urn:nasa:pds:mars2020\_mastcamz\_ops\_cali\linebreak brated:data:ZRF\_0079\_0673952814\_1\linebreak13RAD\_N0032430ZCAM03128\_110085J03}
\end{hangparas}

\bibliographystyle{IEEEtran}
{\normalsize\bibliography{main}}

\end{multicols}

\end{document}